\documentclass[10pt,a4paper,onecolumn]{article}

\usepackage[a4paper,total={6.3in,9.75in}]{geometry}
\usepackage[T1]{fontenc}
\usepackage{palatino}
\usepackage{titling}
\usepackage{graphicx}
\graphicspath{{figures/}}
\usepackage{amsmath,amssymb}
\usepackage{booktabs}
\usepackage{algorithm}
\usepackage{algpseudocode}
\usepackage[numbers]{natbib}
\usepackage{hyperref}
\usepackage{subcaption}
\usepackage{xcolor}
\usepackage{url}
\definecolor{linkblue}{RGB}{0,90,180}

\hypersetup{
  colorlinks=true,
  citecolor=linkblue,
  linkcolor=linkblue,
  urlcolor=linkblue
}

\usepackage{array}
\usepackage{placeins}

\title{Spatially Conditioned Trajectory Diffusion}

\newcommand{\equalSupervisionMark}{\textsuperscript{\fnsymbol{footnote}}}

\author{
Yefim Nudelman, Dor Atzmon\thanks{Equal supervision.}, and Tom Tirer\equalSupervisionMark\\[0.5em]
{\small\itshape Faculty of Engineering, Bar-Ilan University}\\
{\small\itshape Ramat Gan, Israel}
}

\date{}

\begin{document}

\renewcommand{\thefootnote}{\fnsymbol{footnote}}
\maketitle
\renewcommand{\thefootnote}{\arabic{footnote}}

\begin{abstract}
Multi-Robot Motion Planning (MRMP) requires trajectories that avoid environmental obstacles and collisions between robots in a shared workspace. Diffusion-based planners often address obstacle avoidance through sampling-time guidance, while some conditioned approaches aggregate map geometry into a global representation. We present Spatially Conditioned Trajectory Diffusion (SCTD), a map-conditioned trajectory generator for multi-robot planning. SCTD retains multiple position-aware occupancy tokens accessed through cross-attention throughout a temporal U-Net, with hard endpoint inpainting enforcing start and goal states. One model trained on single-robot demonstrations across procedural environments generates obstacle-aware trajectories without sampling-time obstacle-cost guidance. High-level Conflict-Based Search (CBS) coordinates these trajectories through constrained replanning, retaining inter-agent guidance during sampling and reusing the model across workspace layouts and team sizes. Experiments assess single-robot success and trajectory quality, generalization to unfamiliar layouts with added obstacles, and coordinated multi-robot planning. On 15,000 held-out tasks, SCTD achieves higher success than the evaluated guided diffusion baseline, with the largest gains on hard-detour tasks, while trajectory-quality comparisons show smoother trajectories. Conditioning ablations show higher success with spatial tokens than with a global map embedding, and map-replacement tests demonstrate dependence on the supplied geometry. The resulting multi-robot system achieves higher team success than the compared diffusion baselines, with experiments covering teams of up to 15 robots.
\end{abstract}

\par\smallskip

\section{Introduction}

{\emph{Multi-Robot Motion Planning} (MRMP) is a fundamental challenge in robotics,
requiring teams of robots to move through a shared continuous workspace while avoiding
environmental obstacles and one another. Applications such as automated warehouses require
individual routes to account for workspace geometry while remaining compatible with the motions
of other robots. This coupling makes planning computationally challenging
as the team size and complexity of the workspace increase. Obstacle layouts also vary across tasks, requiring planners to adapt their routes to changing workspace geometry.} %
{One approach to this complexity is to formulate a discrete abstraction of MRMP
as \emph{Multi-Agent Path Finding} (MAPF), representing configurations and transitions on a graph,
typically in discrete timesteps ~\cite{stern2019mapf}. MAPF methods such as \emph{Conflict-Based Search} (CBS) separate coordination from individual path planning: a high-level search resolves conflicts by
imposing constraints, while low-level planners revise affected agents' paths~\cite{sharon2015cbs}.
Within this decomposition, coordination repeatedly requests individual paths under updated
constraints. Extending this structure to continuous-space planning therefore motivates studying
low-level generators that account for the current workspace geometry while supplying trajectories
for individual sampling and replanning~\cite{shaoul2025mmd}.}

{Diffusion models \cite{sohl2015deep,song2019generative,ho2020denoising,song2021scorebased}  offer a learned approach to low-level trajectory generation
by modeling distributions over complete trajectories and generating samples through iterative
denoising~\cite{janner2022diffuser}. \emph{Motion Planning Diffusion}~\cite{carvalho2025mpd} (MPD)
develops this approach for robot motion planning, learning trajectory priors
from demonstrations and adapting generated motions through sampling-time
cost-gradient guidance.}

\emph{Multi-robot Multi-model planning Diffusion}
(MMD)~\cite{shaoul2025mmd} combines single-robot diffusion planners
with constraint-based coordination, resolving conflicts through
guided individual replanning. Simultaneous MRMP Diffusion
(SMD)~\cite{liang2025smd} instead plans trajectories for all robots
simultaneously, using constrained optimization during denoising
to address obstacle and inter-robot collisions. MMD's local priors capture motion
patterns associated with their training environments, supporting data-adherent planning. During
sampling, obstacle-cost gradients adapt generated trajectories to workspace geometry, while
inter-robot guidance addresses coordination constraints. The current map therefore influences
sampling, but does not explicitly condition the learned denoiser. This distinction matters when
coordination repeatedly requests trajectories: workspace-dependent routing can also be represented
in the learned prior used for each request. Supplying geometry directly to the denoiser offers this
complementary role, while sampling-time guidance continues to address coordination constraints.

{Explicit environment conditioning offers a complementary way to incorporate
workspace geometry into trajectory generation. \citet{carvalho2022conditioned} encode an occupancy
map and combine its embedding with task features  as a single ``token'' for conditioning a temporal U-Net through
cross-attention. Building on this approach, we examine how trajectory features access the map
during denoising. An aggregated map embedding can encode geometry, while spatially indexed
features presented as separate attention tokens allow different workspace regions to be queried
individually. Spatial map tokens also appear in Roken's joint multi-robot diffusion
transformer~\cite{bai2026roken}; our focus is their use in a single-robot backbone learned across
workspace layouts and evaluated within inherited multi-robot coordination.}

{In this paper, we present \emph{Spatially Conditioned Trajectory Diffusion} (SCTD), a
single-robot trajectory backbone trained across procedurally generated
environments. A convolutional occupancy encoder produces position-aware spatial features that
trajectory features can query separately through cross-attention throughout a temporal U-Net.
Start and goal states remain fixed through hard endpoint inpainting during denoising.
For multi-robot planning, we integrate SCTD as the low-level trajectory
generator in XECBS~\cite{shaoul2025mmd}, retaining inter-agent guidance
to address coordination constraints during sampling and replanning.
This formulation uses one map-conditioned backbone across workspace layouts while
retaining the established coordination machinery.}

{Our contributions are:}
\begin{enumerate}
 
  \item {\textbf{Spatially Conditioned Trajectory Diffusion.}
  We introduce SCTD, a trajectory diffusion model that retains occupancy-derived convolutional
  features as position-aware spatial tokens and exposes them to trajectory features through
  cross-attention throughout a temporal U-Net.}

  \item {\textbf{Generalization and map-dependent generation in unseen workspaces.}
  We show that an SCTD model generates feasible trajectories in
  {held-out maps} without
  static-obstacle-cost guidance, including obstacle-dependent and hard-detour tasks. Compared with MPD, SCTD achieves higher single-agent success.}

\item {\textbf{Reusable low-level planning for multi-robot coordination.}
We integrate SCTD as the low-level trajectory generator in
XECBS~\cite{shaoul2025mmd}, retaining inter-agent guidance and reusing one
map-conditioned model across robots and workspace layouts. On held-out maps,
the resulting system achieves higher team success than MMD at every evaluated
team size from 3 to 15 robots, and than SMD~\cite{liang2025smd} at the shared sizes of 3, 6,
and 9 robots, under a 180-second trial limit.}
\end{enumerate}

\section{Related Work}

\paragraph{Diffusion models for motion planning.}

{Diffuser~\cite{janner2022diffuser} formulates planning as trajectory denoising and combines guidance
for planning objectives with inpainting for fixed trajectory states.
\emph{Motion Planning Diffusion} (MPD)~\cite{carvalho2025mpd} uses learned trajectory priors and adapts samples through
task-specific cost gradients, while EDMP~\cite{saha2024edmp} combines multiple inference-time
{cost objectives. Sampling-time objectives are distinct from
denoiser inputs and task conditions such as endpoint inpainting.}
Environment conditioning and sampling-time cost guidance are not mutually exclusive.
SceneDiffuser~\cite{huang2023scenediffuser} combines scene-conditioned generation with differentiable physical objectives and
goal-directed planning during sampling.
PRESTO~\cite{seo2025presto} conditions its trajectory model on a key-configuration environment representation and examines additional collision and smoothness cost guidance in a sampling
ablation. Learned conditioning supplies an environment-dependent trajectory
prior, while cost guidance adjusts samples according to further objectives/constraints.
Within this trajectory-prior formulation, SCTD focuses on learning workspace-dependent routing
through map conditioning.}

\paragraph{Environment-Conditioned Trajectory Diffusion.}

{Environment-con\-di\-tioned trajectory diffusion differs in the geometric
representations exposed to the denoiser. \citet{carvalho2022conditioned} pretrain an
occupancy-image encoder through signed-distance regression, aggregate its features into a global
map embedding, and combine this embedding with task features to condition a temporal U-Net
through cross-attention. CAMPD~\cite{sandra2026campd} instead attends to a variable set of
structured contextual elements, including obstacle parameters, and uses classifier-free guidance
to strengthen conditioning rather than applying obstacle-cost gradients.
Observation-conditioned approaches include RobotDiffuse~\cite{zhang2024robotdiffuse}, which
encodes obstacle point clouds for transformer-based trajectory generation, and
DiffusionSeeder~\cite{huang2025diffusionseeder}, which uses depth observations to generate
trajectory seeds for subsequent motion optimization. Spatial map tokens are also used in
Roken~\cite{bai2026roken}, where robot tokens cross-attend to position-aware map patches within
a joint multi-robot diffusion transformer. These approaches distinguish aggregated map context,
structured environmental elements, and separately addressable spatial features.
{SCTD uses occupancy-derived spatial features to learn a single-robot trajectory
backbone across procedural environments, rather than a joint team-trajectory model. We assess
this backbone both in single-agent generation and within inherited multi-robot coordination;
the contribution is the resulting planning system and its evaluated use, not spatial tokens or
cross-attention in isolation.}}

\paragraph{Diffusion-based Multi-Robot Motion Planning.}

{Diffusion-based multi-robot planners differ in how trajectory generation is
coupled to coordination. MMD~\cite{shaoul2025mmd} pairs reusable single-robot diffusion planners
with MAPF-inspired coordination, including CBS~\cite{sharon2015cbs}, using only single-robot
training data rather than learning a joint trajectory distribution over the team. Its planning
framework translates spatiotemporal constraints into sampling-time guidance and replans affected
robots under those constraints. Other formulations include joint trajectory generation with
constrained projection in Simultaneous MRMP Diffusion~\cite{liang2025smd} and robot-token-based
joint generation for safety and connectivity in Roken~\cite{bai2026roken}.
Discrete-Guided Diffusion~\cite{liang2026dgd} instead uses MAPF-derived spatiotemporal structure
to guide continuous trajectory generation, followed by constraint repair.
Within MMD's inherited constraint-based coordination framework, we evaluate SCTD's shared
spatially map-conditioned backbone on held-out maps, retaining inter-agent guidance while
assessing its effectiveness as the low-level generator across workspace layouts.}

\section{Preliminaries and Problem Formulation}
\label{sec:preliminaries}

\subsection{Multi-Robot Motion Planning}

Multi-Robot Motion Planning (MRMP) concerns the coordinated motion of
multiple robots in a shared continuous workspace.

An environment $E$ consists of a bounded planar workspace
$\Omega\subset\mathbb{R}^2$ and an obstacle region
$\mathcal{O}_E\subset\Omega$. The environment is represented by a
binary occupancy grid $M_E$, with occupied cells marked by one.

We consider $N$ disk robots of radius $r>0$. At waypoint $h$, robot
$i$ has state $s_{i,h}=(q_{i,h},v_{i,h})\in\mathbb{R}^4$, where
$q_{i,h}\in\mathbb{R}^2$ denotes its planar position and
$v_{i,h}\in\mathbb{R}^2$ contains its velocity channels.
Its trajectory is represented by a sequence of $H_{\mathrm{traj}}$
states:
\begin{equation}
  \tau_i=(s_{i,0},\ldots,s_{i,H_{\mathrm{traj}}-1}).
\end{equation}

Together, these trajectories form a joint plan
$\boldsymbol{\tau}=(\tau_1,\ldots,\tau_N)$. Given prescribed start
and goal states $s_i^{\mathrm{s}}$ and $s_i^{\mathrm{g}}$, MRMP seeks
a joint plan satisfying
\begin{equation}
\begin{aligned}
  s_{i,0} &= s_i^{\mathrm{s}},
  && i=1,\ldots,N,\\
  s_{i,H_{\mathrm{traj}}-1} &= s_i^{\mathrm{g}},
  && i=1,\ldots,N,\\
  \mathbf{q}_h &\in \mathcal{Q}_{\mathrm{free}}(E),
  && h=0,\ldots,H_{\mathrm{traj}}-1.
\end{aligned}
\end{equation}
Here, $\mathbf{q}_h=(q_{1,h},\ldots,q_{N,h})$ denotes the joint
position at synchronized waypoint $h$. The set
$\mathcal{Q}_{\mathrm{free}}(E)$ contains joint configurations in
which robot centers lie within the workspace and robot footprints
intersect neither obstacles nor one another.
\subsection{{MAPF and Conflict-Based Coordination}}

Multi-Agent Path Finding (MAPF) formulates coordinated routing on a graph, typically using
discrete timesteps~\cite{stern2019mapf}. Conflict-Based Search (CBS) separates this problem
into high-level coordination and low-level individual planning~\cite{sharon2015cbs}.
A high-level search node contains individual paths and their constraints. When two paths
conflict, the search branches by imposing an additional constraint on either agent and
replanning the affected path.

This decomposition provides a framework for coordinating continuous trajectory generators:
individual planning produces candidate motions, while the coordination layer identifies
incompatibilities and requests replanning under additional conditions~\cite{shaoul2025mmd}.
The decomposition can be retained while replacing graph-based individual planning with
continuous-space trajectory generation.

The proposed method specifies how SCTD serves as the individual trajectory generator
within the selected coordination procedure.

\subsection{{Diffusion Models}}

A denoising diffusion probabilistic model (DDPM)~\cite{ho2020denoising} learns to generate data by
reversing a gradual noising process. For trajectory
generation, let $u^0$ denote a clean trajectory representation~\cite{janner2022diffuser}.
At a sampled diffusion step $k$, the forward process adds Gaussian noise:
\begin{equation}
  u^k=\sqrt{\bar{\alpha}_k}\,u^0
  +\sqrt{1-\bar{\alpha}_k}\,\epsilon,
  \qquad \epsilon\sim\mathcal{N}(0,I),
\end{equation}
where $\bar{\alpha}_k$ is determined by the noise schedule. The denoiser learns
to predict the added noise from the corrupted trajectory and the diffusion step.
At inference, generation starts from Gaussian noise and applies successive
reverse updates, $u^K\rightarrow\cdots\rightarrow u^0$, where $K$ is the number
of diffusion steps. Each update revises the complete trajectory representation
rather than generating waypoints sequentially.

\subsection{{Conditional Trajectory Diffusion}}

In conditional diffusion, the denoiser also receives context $c$:
\begin{equation}
  \widehat{\epsilon}^{\,k}=\epsilon_\theta(u^k,k,c).
\end{equation}
The context guides the denoising process toward samples compatible with the
specified task or environment.

In SCTD, the environmental context is the encoded occupancy map,
$c=Z_E=E_\phi(M_E)$. Start and goal states are handled separately: endpoint
inpainting fixes them before denoising and restores them after reverse updates.
Thus, map conditioning informs generation of the trajectory interior, whereas
inpainting enforces its boundary states. Section~\ref{sec:method} describes the
spatial conditioning interface and its integration with multi-robot coordination.

\section{Spatially Conditioned Trajectory Diffusion}
\label{sec:method}

In this section, we present SCTD from architecture to multi-robot integration. We first present
the spatial-conditioning architecture, then the training and single-robot inference procedures,
and finally how the shared generator provides per-robot trajectories within MRMP coordination.

\subsection{SCTD Architecture}

Building on the conditional denoising formulation (Section~\ref{sec:preliminaries}), SCTD takes
an occupancy map and prescribed start and goal states as input. It encodes the map as spatial
tokens and predicts noise for a trajectory conditioned on those tokens; endpoint inpainting fixes
the boundary states, and reverse diffusion produces the complete trajectory.

\begin{figure*}[t]
  \centering
  \includegraphics[width=0.92\linewidth]{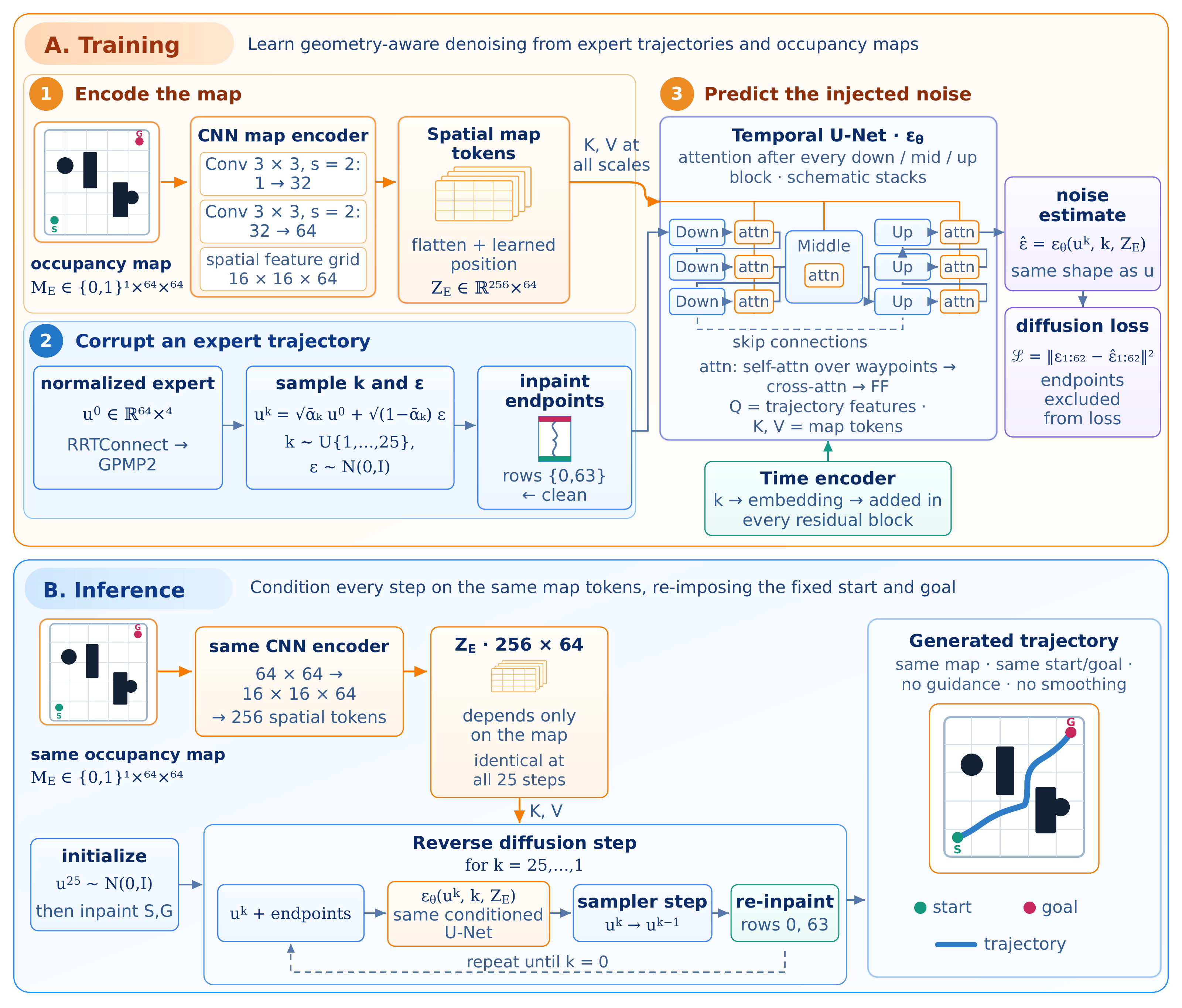}
  \caption{Training and inference of the map-conditioned trajectory diffuser. During training,
  the temporal U-Net predicts noise added to normalized expert trajectories, conditioned on
  spatial occupancy-map tokens and the diffusion timestep. Endpoints are fixed by inpainting
  and excluded from the loss. During single-agent inference, map tokens depend only on the map and
  are therefore identical at every reverse step; endpoints are restored after every update. The
  raw trajectory is
  converted back to world coordinates without static-obstacle-cost guidance or a separate smoother;
  this describes the single-agent pipeline, not the multi-agent system. Start/goal markers on the
  maps are illustrative only.}
  \label{fig:architecture}
\end{figure*}

\paragraph{Spatial map representation.}
As illustrated in Figure~\ref{fig:architecture}, a convolutional encoder
maps the binary occupancy grid $M_E$ to a
$16\times16\times64$ feature grid $F_{\phi_{\mathrm{s}}}(M_E)$.
Flattening its spatial dimensions and adding learned positional embeddings
produces the map tokens:
\begin{equation}
  Z_E=E_\phi(M_E)
  =\operatorname{Flatten}\!\left(F_{\phi_{\mathrm{s}}}(M_E)\right)+P,
  \qquad Z_E,P\in\mathbb{R}^{256\times64},
  \label{eq:complete-map-encoder}
\end{equation}
where $\phi=(\phi_{\mathrm{s}},P)$ collects the map encoder's trainable parameters:
$\phi_{\mathrm{s}}$ denotes the convolutional stem and $P$ the learned positional embeddings.
Each row of $P$ corresponds to a fixed grid location. Each token
therefore combines an occupancy-derived feature with positional
information. Keeping the tokens separate allows trajectory features to query
different map regions individually. Analytic signed-distance values supervise
stem pretraining, as described in the Training subsection; occupancy is the
denoiser's environmental input.

\paragraph{Map-conditioned trajectory denoiser.}
A temporal U-Net processes the noisy trajectory
$u^k\in\mathbb{R}^{64\times4}$ at diffusion step $k$ and predicts
the added noise:
\begin{equation}
  \widehat{\epsilon}^{\,k}
  =\epsilon_\theta(u^k,k,Z_E),
  \qquad \widehat{\epsilon}^{\,k}\in\mathbb{R}^{64\times4}.
  \label{eq:conditional-noise-prediction}
\end{equation}
The discrete diffusion step $k$ is encoded with sinusoidal features and an MLP; in every
residual convolutional block, a learned projection of this embedding is added channel-wise
to the intermediate features. Each down, middle, and up stage of the U-Net combines residual
convolutional blocks with one transformer layer, which applies self-attention over trajectory
waypoints, cross-attention to the map tokens, and a feed-forward layer, each with a residual
connection. In cross-attention, trajectory features provide the queries and $Z_E$ provides
the keys and values, so the map tokens are the only conditioning context. Start and goal
states are fixed separately through hard endpoint inpainting. Detailed layer dimensions and
configuration are provided in Appendix~\ref{sec:implementation-details}.

\subsection{Training}

\paragraph{Occupancy-encoder pretraining.}
We first initialize the convolutional occupancy stem through
signed-distance regression, motivated by the encoder pretraining
approach of Carvalho et al.~\cite{carvalho2022conditioned}.
Our pretraining head operates on the spatial feature grid used to
construct SCTD's map tokens. For a sampled workspace coordinate $x$,
bilinear interpolation extracts a local feature vector from this grid.
A lightweight MLP receives this feature together with $x$ and predicts
the clipped analytic obstacle signed distance $\bar d_E(x)$ at query
points drawn uniformly over the workspace.
The head reads only the stem's output features, without connections
to intermediate encoder features. The signed distance is positive
in free space, zero on obstacle boundaries, and negative inside
obstacles.

Let $\phi_{\mathrm{s}}$ denote the occupancy-stem parameters within
$\phi$, and $\xi$ the query-head parameters. We optimize
\begin{equation}
  \mathcal{L}_{\mathrm{SDF}}(\phi_{\mathrm{s}},\xi)
  =
  \mathbb{E}_{M_E,x}
  \left[
    \left|
      \bar d_E(x)
      -\widehat{d}_{\phi_{\mathrm{s}},\xi}(M_E,x)
    \right|
  \right].
  \label{eq:sdf-pretraining}
\end{equation}
This stage uses occupancy maps and signed-distance targets without
trajectory demonstrations. The query head is then discarded, and
the pretrained stem initializes the map encoder $E_\phi$.

\paragraph{Joint encoder and trajectory-diffusion training.}
We next optimize the map encoder and temporal denoiser jointly on
occupancy maps and expert trajectories. The pretrained stem remains
trainable; the positional embeddings and the entire temporal U-Net,
including its cross-attention layers, are trained from random
initialization in this stage.
For a clean expert trajectory $u^0$, we sample a diffusion step $k$
uniformly and noise $\epsilon\sim\mathcal{N}(0,I)$, and construct
$u^k$ using the forward process in Section~\ref{sec:preliminaries}.

Let $u^{\mathrm{s}}$ and $u^{\mathrm{g}}$ denote the prescribed start
and goal states in the trajectory representation.
The inpainting operator
$\mathcal{I}_{u^{\mathrm{s}},u^{\mathrm{g}}}$ replaces trajectory
rows 0 and 63 with these states while leaving interior rows unchanged.
The noise-prediction objective is
\begin{equation}
  \mathcal{L}_{\mathrm{diff}}(\theta,\phi)
  =
  \mathbb{E}_{u^0,M_E,k,\epsilon}
  \left[
    \left\|
      \left[
        \epsilon
        -
        \epsilon_\theta\!\left(
          \mathcal{I}_{u^{\mathrm{s}},u^{\mathrm{g}}}(u^k),
          k,E_\phi(M_E)
        \right)
      \right]_{1:62}
    \right\|_2^2
  \right],
  \label{eq:conditional-diffusion-loss}
\end{equation}
where $[\cdot]_{1:62}$ selects interior rows 1 through 62.
The denoiser therefore receives clean endpoints at every sampled
noise level, and endpoint prediction entries are overwritten before
the loss, so they contribute no gradient;
Eq.~\eqref{eq:conditional-diffusion-loss} gives the equivalent
objective. Map context is supplied through the occupancy-derived
tokens defined in the Architecture subsection. Signed-distance
regression is used only for initialization; subsequent joint
training optimizes the diffusion objective.

Algorithm~\ref{alg:sctd-training} summarizes the two-stage training procedure; Appendix~\ref{sec:implementation-details} provides implementation-specific details.

\begin{algorithm}[t]
\small
\caption{SCTD training}
\label{alg:sctd-training}
\begin{algorithmic}[1]
\Statex \textbf{Input:} SDF pretraining data $\mathcal{D}_{\mathrm{SDF}}$, trajectory demonstrations
         $\mathcal{D}_{\tau}$
\Statex \textbf{Output:} map encoder $E_\phi$ and trajectory denoiser $\epsilon_\theta$
\Statex \textit{Stage 1: occupancy-stem pretraining}
\While{the pretraining criterion is not reached}
  \State Sample $(M_E,x,\bar d_E(x))\sim\mathcal{D}_{\mathrm{SDF}}$
  \State $\widehat d\gets\widehat d_{\phi_{\mathrm{s}},\xi}(M_E,x)$
  \State $\mathcal{L}_{\mathrm{SDF}}\gets|\bar d_E(x)-\widehat d|$
  \State Update $\phi_{\mathrm{s}}$ and $\xi$ using $\nabla\mathcal{L}_{\mathrm{SDF}}$
\EndWhile
\State Discard the SDF query head
\Statex \textit{Stage 2: joint trajectory-diffusion training}
\While{the diffusion-training criterion is not reached}
  \State Sample $(M_E,u^0,u^{\mathrm{s}},u^{\mathrm{g}})\sim\mathcal{D}_{\tau}$
  \State Sample $k\sim\mathcal{U}\{1,\ldots,K\}$ and $\epsilon\sim\mathcal{N}(0,I)$
  \State $u^k\gets\sqrt{\bar\alpha_k}u^0+\sqrt{1-\bar\alpha_k}\epsilon$
  \State $u^k\gets\mathcal{I}_{u^{\mathrm{s}},u^{\mathrm{g}}}(u^k)$
  \State $Z_E\gets E_\phi(M_E)$
  \State $\widehat{\epsilon}^{\,k}\gets\epsilon_\theta(u^k,k,Z_E)$
  \State $\mathcal{L}_{\mathrm{diff}}\gets
         \left\|[\epsilon-\widehat{\epsilon}^{\,k}]_{1:62}\right\|_2^2$
  \State Update $\phi$ and $\theta$ using $\nabla\mathcal{L}_{\mathrm{diff}}$
\EndWhile
\State \Return $E_\phi,\epsilon_\theta$
\end{algorithmic}
\end{algorithm}

\begin{figure*}[!tp]
  \centering
  \includegraphics[width=0.96\linewidth]{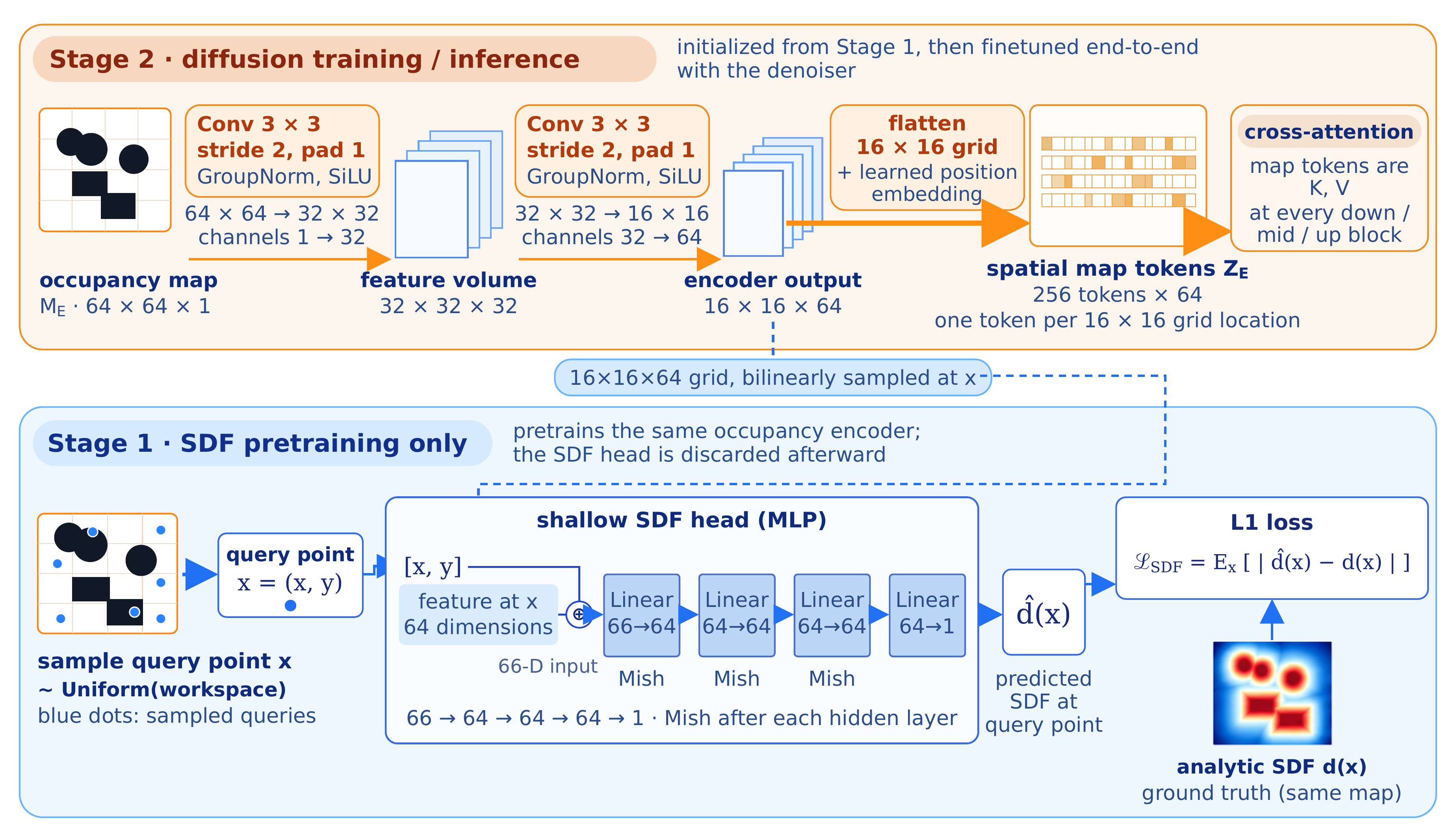}
  \caption{The occupancy encoder $E_\phi$ and its two-stage training. A binary occupancy map is
  converted into 256 position-aware, 64-dimensional tokens. These tokens condition the denoiser
  through cross-attention and retain spatial structure instead of compressing the map into a
  single global vector. In Stage 1 (bottom), the convolutional stem is pretrained to predict clipped
  analytic signed distances at sampled workspace points using an auxiliary head and an $L^1$ loss.
  The head is then discarded, and the stem is finetuned jointly with the denoiser in Stage 2
  (top). Analytic SDF supplies supervision only; occupancy is the planner's only learned
  environmental input. Token and SDF illustrations are schematic; $\oplus$ denotes concatenation.
  Additional head and sampling details appear in Appendix~\ref{sec:implementation-details}.}
  \label{fig:vision-encoder}
\end{figure*}

Figure~\ref{fig:vision-encoder} details the occupancy encoder and its two-stage training.

\subsection{Inference}

Given an occupancy map and prescribed start and goal states, SCTD
generates a trajectory by iteratively denoising an initial Gaussian
sample. At each reverse step, the denoiser predicts noise conditioned
on the spatial map tokens and the diffusion step, and uses it to compute $u^{k-1}$:
\begin{equation}
  u^{k-1}=\frac{1}{\sqrt{\alpha_k}}\Big(u^k-\frac{\beta_k}{\sqrt{1-\bar\alpha_k}}\,
  \epsilon_\theta(u^k,k,Z_E)\Big)+s\,\sigma_k z,
  \qquad z\sim\mathcal{N}(0,I),
  \label{eq:reverse-step}
\end{equation}
where $\alpha_k=1-\beta_k$, $\sigma_k^2=\frac{1-\bar\alpha_{k-1}}{1-\bar\alpha_k}\beta_k$, $s=0.5$ scales the sampling noise
as in MMD~\cite{shaoul2025mmd}, and $z=0$ at the final step.
Equation~\eqref{eq:reverse-step} gives the update without clipping; in implementation, the predicted clean trajectory is clipped to $[-1,1]$ before computing the reverse mean, as specified in Table~\ref{tab:config}.
The inpainting
operator defined in Training imposes the endpoints before denoising
and restores them after every update. This preserves the boundary
states throughout sampling and matches the endpoint conditioning
used during training. Alg.~\ref{alg:sctd-inference} summarizes
the single-agent procedure.

Note that during inference, obstacle information enters through the
learned map representation, sparing the need for sampling-time static-obstacle-cost
guidance or a separate trajectory smoother. Repeated sampling provides
multiple candidate trajectories for the same task. Collision-free
feasibility is assessed separately using the evaluation protocol
in Section~\ref{sec:experiments}. The multi-agent integration described
next adds coordination constraints to this single-robot generation
process. 
\begin{algorithm}[t]
\small
\caption{Single-agent inference}
\label{alg:sctd-inference}
\begin{algorithmic}[1]
\Statex \textbf{Input:} occupancy map $M_E$, endpoint states $u^{\mathrm{s}},u^{\mathrm{g}}$,
         trained map encoder $E_\phi$ and denoiser $\epsilon_\theta$
\Statex \textbf{Output:} world-space trajectory $\tau$
\State $Z_E\gets E_\phi(M_E)$
\State Let $K$ denote the number of reverse diffusion steps
\State Sample $u^K\sim\mathcal{N}(0,I)$ and impose endpoints $u^{\mathrm{s}},u^{\mathrm{g}}$
\For{$k=K,K-1,\ldots,1$}
  \State Predict noise $\widehat{\epsilon}^{\,k}\gets\epsilon_\theta(u^k,k,Z_E)$
  \State Update $u^k$ to $u^{k-1}$ with the reverse diffusion step
  \State Impose endpoints $u^{\mathrm{s}},u^{\mathrm{g}}$ on $u^{k-1}$
\EndFor
\State Convert $u^0$ to a world-space trajectory $\tau$
\State \Return $\tau$
\end{algorithmic}
\end{algorithm}

\subsection{SCTD-MRMP Integration}
\label{sec:sctd-multiagent}

For multi-robot planning, we reuse the same SCTD model to generate
trajectories for every robot, both during initial planning and when
conflicts require replanning. 
Each planning request supplies the workspace occupancy
map and the requesting robot's prescribed endpoints. We coordinate
these trajectories using XECBS~\cite{shaoul2025mmd}, a variant of CBS ~\cite{sharon2015cbs} which detects
inter-agent conflicts at synchronized waypoints, branches on additional
constraints, and replans the affected robot. Search continues until
the selected trajectories have no detected inter-agent conflicts,
or the planner fails or reaches its runtime limit.

Coordination constraints influence trajectory generation through
sampling-time guidance. The denoiser's learned context remains
map-only: spatial tokens supply workspace information, while
constraint guidance accounts for other robots' current trajectories.

The high-level SCTD with XECBS procedure used in our multi-robot evaluation is summarized in Algorithm~\ref{alg:sctd-xecbs}.

\section{Experimental Results}
\label{sec:experiments}

Our experiments address three questions: how reliably SCTD generates feasible trajectories on
unseen maps, whether its outputs genuinely depend on the supplied map,
and how effectively the same generator supports multi-robot coordination.
Bold values indicate the best result within each comparison, including ties at the reported precision.

\subsection{Experimental Setup}

\paragraph{Environment and data.}
Experiments use disk robots of radius $r=0.05$ in $\Omega=[-1,1]^2$, with disk and axis-aligned
rectangular obstacles. SCTD observes a $64\times64$ binary occupancy grid, while task construction
and feasibility evaluation use analytic geometry. The principal single-agent benchmark contains
15,000 tasks on 500 maps held out from training.

\paragraph{Metrics.}
A single-agent trajectory is feasible if the robot avoids obstacles and remains within the
workspace at every waypoint and interpolated check point. Success@$J$ (or simply @$J$) is the percentage of tasks
with at least one feasible trajectory among $J$ samples, computed separately for each map and
averaged equally across maps. Team success is the percentage of multi-agent tasks solved within
the trial time limit, requiring feasible trajectories for all robots and no detected inter-agent
conflicts.

\subsection{Single-Agent Performance on Unseen Maps}

\paragraph{Single-agent baselines.}
We compare SCTD with MPD using MMD's public
implementation~\cite{shaoul2025mmd}. MPD applies SDF-based obstacle
guidance and post-sampling smoothing, whereas SCTD uses occupancy
conditioning without obstacle-cost guidance or a separate smoother.
Both methods use the same tasks and feasibility criterion.

Table~\ref{tab:sealed-success} reports single-agent success overall and on obstacle-dependent tasks, split into mild and hard-detour tasks (defined in Appendix~\ref{sec:evaluation-protocol}).
SCTD achieves higher success than MPD overall, both with one sample (90.5\% versus 82.3\%)
and with 64 samples (99.2\% versus 89.4\%). The largest differences occur on
obstacle-dependent tasks, particularly hard detours, while both diffusion planners
perform near ceiling on straight-clear tasks (not shown).

\begin{table}[t]
  \caption{Single-agent success (\%) on 15,000 held-out tasks across 500 maps, averaged
  equally across maps. Success@$J$ measures whether at least one of $J$ candidates
  is feasible.}
  \label{tab:sealed-success}
  \centering
  \small
  \setlength{\tabcolsep}{7pt}
  \renewcommand{\arraystretch}{1.2}
  \begin{tabular}{@{\extracolsep{\fill}}l rr rr}
    \toprule
    Task group & \multicolumn{2}{c}{Success@1 (\%)} &
                 \multicolumn{2}{c}{Success@64 (\%)} \\
    \cmidrule(lr){2-3}\cmidrule(lr){4-5}
    & SCTD & MPD & SCTD & MPD \\
    \midrule
    Overall & \textbf{90.5} & 82.3 & \textbf{99.2} & 89.4 \\
    Mild obstacle-dependent & \textbf{88.6} & 79.3 & \textbf{99.4} & 91.7 \\
    Hard-detour & \textbf{72.7} & 33.8 & \textbf{97.7} & 52.1 \\
    \bottomrule
  \end{tabular}
\end{table}
\paragraph{Single-agent trajectory quality.}

We measure path length and geometric smoothness on feasible trajectories. Our smoothness
measure is the maximum local turn angle, which captures the sharpest change in direction
over a path distance of one robot diameter ($w=0.1$). Smaller angles indicate gentler local
turns. Principal quality comparisons use tasks for which every compared method produces
at least one feasible trajectory. 

We average over feasible samples within each task, then over tasks within each map, and finally equally across maps.
For the trajectory-quality comparison, we also include an

A* planner~\cite{hart1968formal}.

\begin{figure}[t]
  \centering
  \includegraphics[width=0.75\columnwidth]{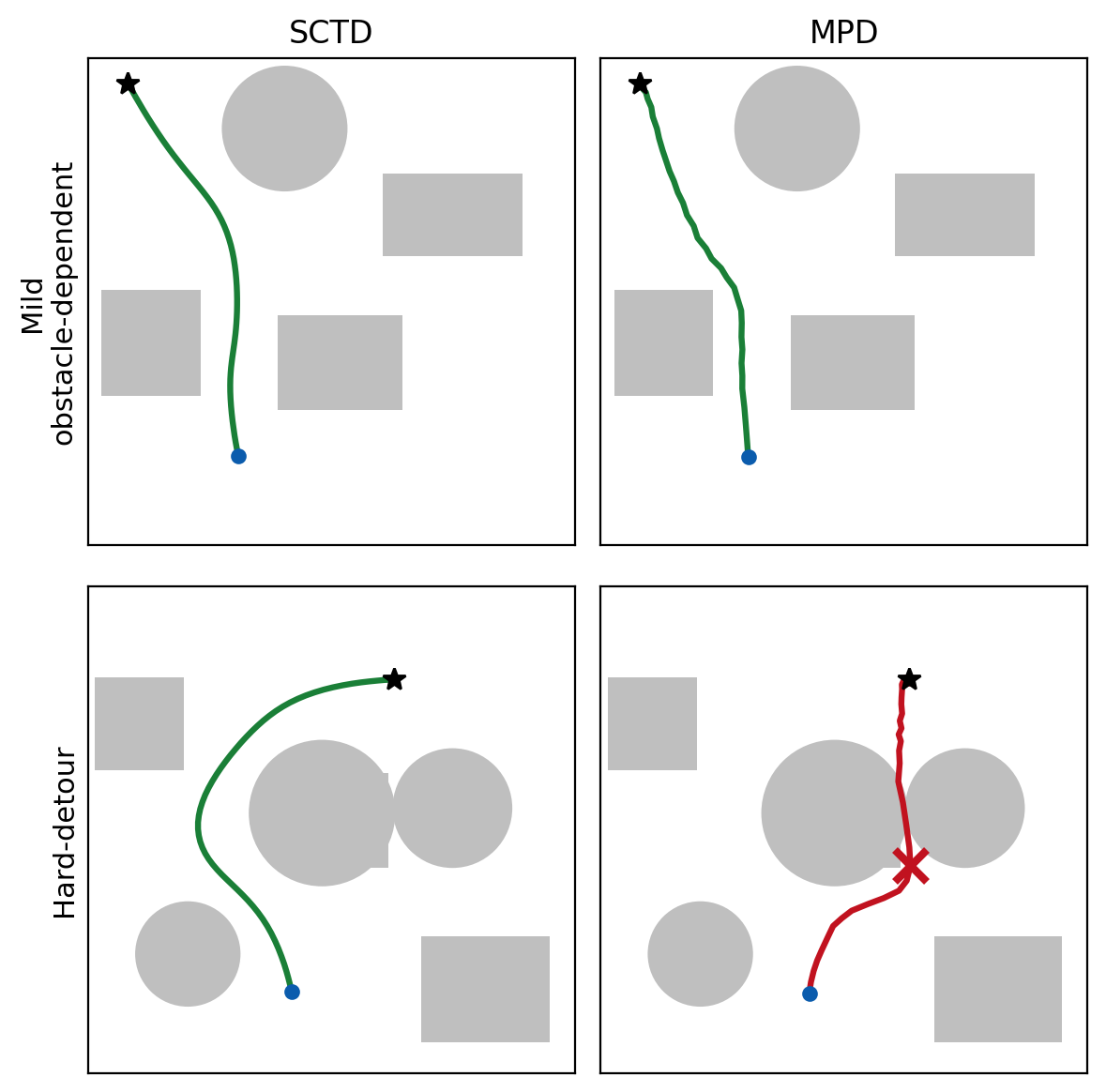}
  \caption{First-sample trajectories from SCTD and MPD on one held-out mild
  obstacle-dependent task (top) and one hard-detour task (bottom). Grey regions denote obstacles; green
  and red paths are feasible and colliding, respectively, and a red cross marks the
  first collision. Blue circles mark starts and black stars mark goals.}

  \label{fig:trajectory-grid}
\end{figure}

As shown in Table~\ref{tab:singleagent-quality}, on commonly solved tasks, SCTD produces shorter paths and smoother paths than the
evaluated A* planner. Compared with MPD, SCTD has
a similar mean path length but markedly smaller maximum local turn angles. Figure~\ref{fig:trajectory-grid} illustrates generated trajectories across two task groups, complementing the population-level success and quality comparisons. 
MPD's large turn angles frequently occur near the trajectory end, as examined further in Appendix~\ref{sec:large-turn-location}.

\begin{table}[t]
  \caption{Single-agent path length and geometric smoothness on tasks with feasible
  outputs. Smoothness is measured by the maximum local turn
  angle over a path distance of one robot diameter; smaller angles indicate gentler
  local turns. Values are means $\pm$ standard deviations across maps.}
  \label{tab:singleagent-quality}
  \centering
  \small
  \setlength{\tabcolsep}{4pt}
  \renewcommand{\arraystretch}{1.15}
  \begin{tabular}{lcc}
    \toprule
    Method & Path length $\downarrow$ &
    \shortstack{Smoothness:\\max.\ turn angle ($^\circ$) $\downarrow$} \\
    \midrule
    SCTD & $\mathbf{1.255} \pm 0.220$ & $\mathbf{17.8} \pm 4.7$ \\
    MPD  & $\mathbf{1.255} \pm 0.170$ & $125.6 \pm 10.1$ \\
    A*   & $1.438 \pm 0.260$ & $80.7 \pm 4.3$ \\
    \bottomrule
  \end{tabular}
\end{table}

\subsection{Map-Pathway Ablation}
\label{sec:map-ablation}

We compare SCTD's spatial map tokens with the global map embedding of
\citet{carvalho2022conditioned}: %
an SDF-pretrained CNN encodes the map as a
128-dimensional vector, which is combined with the normalized start and goal into one
cross-attention token. 

Both designs use the same diffusion-training data and budget, as described in Appendix~\ref{sec:implementation-details}.

For each design, we evaluate a frozen encoder, as in Carvalho et al., and a jointly
fine-tuned encoder, as in SCTD.

All four configurations use the same 500 evaluation maps and 15,000 tasks; this population is held out from training and separate from the sealed test set.
For this ablation, single-sample success (Success@1) is estimated from the fraction of feasible trajectories among the 64 samples per task; Success@64 records whether at least one is feasible. Task-level values are averaged within each map and then equally across maps.

Table~\ref{tab:map-ablation} shows that SCTD outperforms the global embedding across all reported task groups with both frozen and fine-tuned encoders. The advantage is especially pronounced on tasks requiring obstacle avoidance: on hard-detour tasks, the fine-tuned SCTD achieves 73.7\% Success@1 and 97.5\% Success@64, compared with 2.6\% and 38.6\% for the fine-tuned global model. Fine-tuning improves SCTD's success across all groups, whereas it reduces success for the global model. SCTD's advantage also persists when both encoders are frozen, supporting the spatial conditioning design beyond the benefit of encoder fine-tuning alone.

\begin{table}[!t]
  \caption{Map-pathway ablation on the 500-map evaluation population (15,000 tasks), held out from training and drawn from the same generator as the sealed test set. Success (\%) is averaged per map and then equally across maps.}
  \label{tab:map-ablation}
  \centering
  \small
  \setlength{\tabcolsep}{3pt}
  \renewcommand{\arraystretch}{1.1}
  \begin{tabular}{llrrrr}
    \toprule
    Task group & Encoder & \multicolumn{2}{c}{SCTD} & \multicolumn{2}{c}{Global} \\
    \cmidrule(lr){3-4}\cmidrule(lr){5-6}
    & & @1 & @64 & @1 & @64 \\
    \midrule
    Overall & Fine-tuned & \textbf{91.0} & \textbf{99.4} & 43.4 & 73.7 \\
            & Frozen     & \textbf{87.0} & \textbf{98.8} & 50.7 & 80.1 \\
    \midrule
    Mild obstacle-dependent & Fine-tuned & \textbf{89.4} & \textbf{99.4} & 7.1 & 57.6 \\
                            & Frozen     & \textbf{82.1} & \textbf{98.6} & 17.0 & 67.9 \\
    \midrule
    Hard-detour & Fine-tuned & \textbf{73.7} & \textbf{97.5} & 2.6 & 38.6 \\
                & Frozen     & \textbf{65.8} & \textbf{96.2} & 8.2 & 53.0 \\
    \bottomrule
  \end{tabular}
\end{table}

\subsection{Map Dependence and Out-of-Distribution Evaluation}
\label{sec:ood-mmd-envs}

We first test whether SCTD's trajectories depend on the supplied
geometry. Replacing the input map with a wrong map, while keeping the endpoints
fixed, reduces its collision-free rate (on the original map) from 87.3\% to 7.6\% on
obstacle-dependent held-out tasks, supporting dependence on the
supplied obstacle layout. Details are reported in Appendix~\ref{sec:map-dependence-control}.

\begin{figure}[!tp]
  \centering
  \begin{subfigure}{\columnwidth}
    \centering
    \includegraphics[width=0.91\columnwidth,height=0.40\textheight,keepaspectratio]{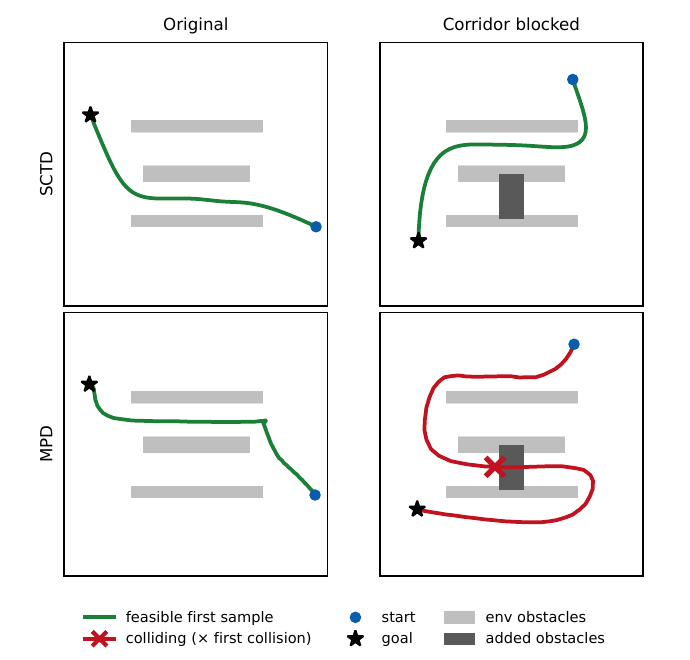}
    \caption{Conveyor environment.}
    \label{fig:ood-conveyor-mapuse}
  \end{subfigure}

  \vspace{0.5em}

  \begin{subfigure}{\columnwidth}
    \centering
    \includegraphics[width=0.92\columnwidth,height=0.40\textheight,keepaspectratio]{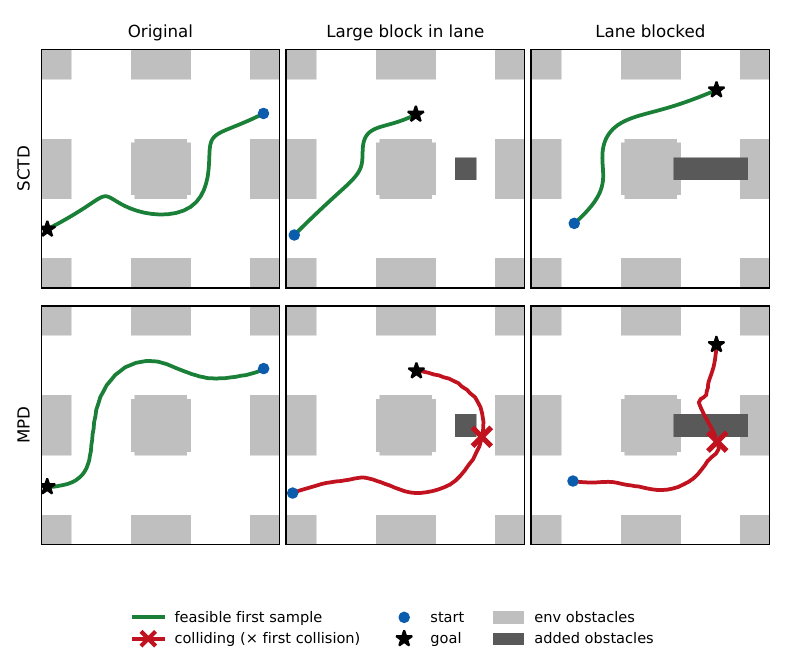}
    \caption{Highways environment.}
    \label{fig:ood-highways-mapuse}
  \end{subfigure}
  \caption{Trajectory examples for MPD and SCTD on selected tasks in MMD's Conveyor (a) and Highways (b) environments. The map-conditional model (SCTD) handles new obstacles better than a model that relies on guidance (MPD).}

  \label{fig:ood-mapuse}
\end{figure}

We then evaluate generalization beyond the procedural training
distribution using MMD's Conveyor and Highways
environments~\cite{shaoul2025mmd}, adding rounded boxes to passages
traversed by the demonstrations. Across 26 maps, we evaluate 5,200 reachability-screened start--goal tasks (200 per map), with 64 ordered samples per task.
MPD uses MMD's released checkpoint for each environment with SDF obstacle guidance and smoothing; SCTD uses raw output without static-obstacle guidance or smoothing and was not trained or tuned on these environments. Here, OOD-$k$ denotes added obstacles absent from MPD's training data. Table~\ref{tab:ood-mmd-envs} shows higher Success@1 and Success@64
for SCTD on the added-obstacle scenes. SCTD achieves at
least 99.5\% Success@64 in every displayed scene, while MPD's success
varies substantially; on Conveyor OOD-4, the rates are 100.0\% and
7.0\%, respectively. Figure~\ref{fig:ood-mapuse} illustrates the corresponding
trajectory examples. Results for all evaluated scenes are reported
in Table~\ref{tab:ood-mmd-envs-full}.

\begin{table}[htb]
  \caption{Single-agent success (\%) on selected added-obstacle scenes in MMD's Conveyor and Highways environments. Each value is averaged equally across maps.
  See Appendix~\ref{app:ood-mmd-envs} for all scenes and their definitions.}
  \label{tab:ood-mmd-envs}
  \centering
  \small
  \setlength{\tabcolsep}{4pt}
  \begin{tabular}{@{}lrrrr@{\hspace{10pt}}lrrrr@{}}
    \toprule
    \multicolumn{5}{c}{Conveyor} & \multicolumn{5}{c}{Highways} \\
    \cmidrule(lr){1-5}\cmidrule(lr){6-10}
    Scene & \multicolumn{2}{c}{MPD} & \multicolumn{2}{c}{SCTD} &
    Scene & \multicolumn{2}{c}{MPD} & \multicolumn{2}{c}{SCTD} \\
    \cmidrule(lr){2-3}\cmidrule(lr){4-5}\cmidrule(lr){7-8}\cmidrule(lr){9-10}
    & @1 & @64 & @1 & @64 & & @1 & @64 & @1 & @64 \\
    \midrule
    OOD-3 & 4.5 & 71.0 & \textbf{62.0} & \textbf{99.5} &
    OOD-3 & 51.6 & 89.4 & \textbf{87.1} & \textbf{100.0} \\
    OOD-4 & 0.0 & 7.0 & \textbf{73.0} & \textbf{100.0} &
    OOD-4 & 44.6 & 49.5 & \textbf{82.4} & \textbf{100.0} \\
    OOD-5 & 47.5 & 98.5 & \textbf{64.0} & \textbf{99.8} &
    & & & & \\
    \bottomrule
  \end{tabular}

  \par\smallskip

\end{table}

\newpage
\subsection{Multi-Agent Results}
\label{sec:multi-agent-results}

\begin{table}[t]
  \caption{Multi-robot team success (\%) within the 180-second
  trial limit. Dashes indicate team sizes not evaluated for SMD.}
  \vspace{-3mm}
  \label{tab:multiagent-success}
  \centering
    \small
    \setlength{\tabcolsep}{6pt}
    \renewcommand{\arraystretch}{1.15}
  \begin{tabular}{crrr}
    \toprule
    Robots & SCTD & MMD & SMD \\
    \midrule
    3  & \textbf{96.7} & 67.8 & 33.3 \\
    6  & \textbf{90.3} & 50.1 & 3.7  \\
    9  & \textbf{82.2} & 33.5 & 0.0  \\
    12 & \textbf{71.3} & 22.1 & --   \\
    15 & \textbf{60.3} & 15.1 & --   \\
    \bottomrule
  \end{tabular}
\end{table}

\begin{figure}[!t]
  \centering
  \includegraphics[width=0.94\columnwidth]{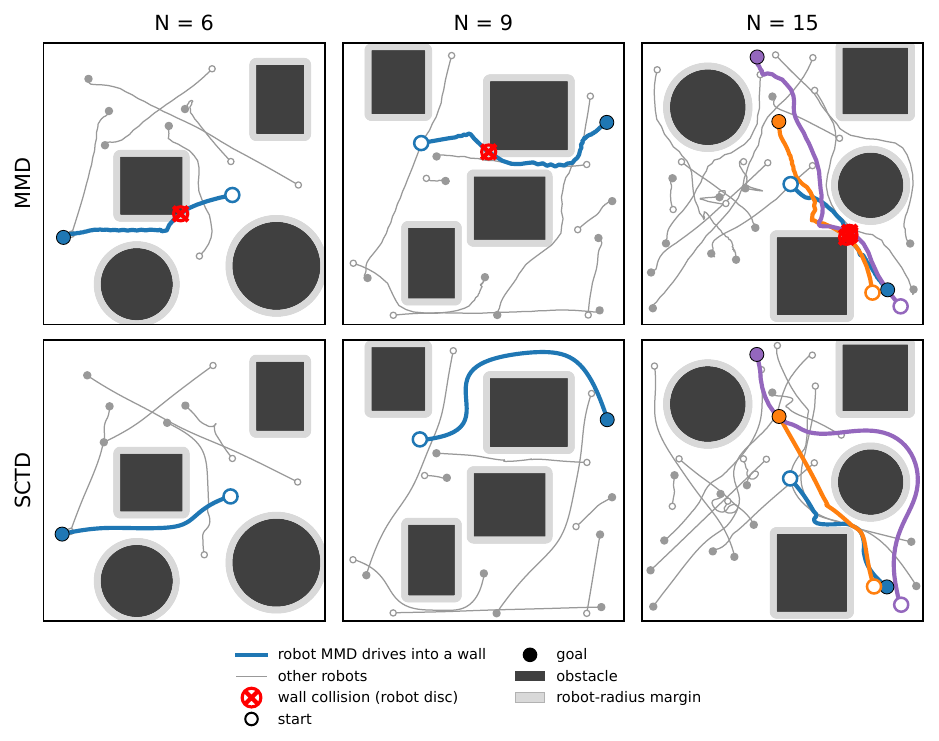}
  \vspace{-2mm}
  \caption{Selected multi-robot trajectory comparisons for teams
  of 6, 9, and 15 robots, with MMD above and SCTD below.
  Colored paths highlight robots whose MMD trajectories collide
  with obstacles; corresponding SCTD paths avoid these collisions.
  Other robots' paths are gray. Red markers indicate obstacle
  collisions accounting for the robot radius, and light-gray
  obstacle margins depict that radius.}
 
  \label{fig:multiagent-win-by-n}
  \vspace{-2mm}
\end{figure}

\paragraph{Multi-agent baselines.}
We evaluate coordinated planning on 100 maps held out from training,
with 10 newly sampled team tasks per map at each evaluated team size. Tasks are shared across
methods. We compare SCTD and MMD using XECBS
coordination~\cite{shaoul2025mmd}, 
and additionally
evaluate SMD~\cite{liang2025smd} for teams of 3, 6, and 9 robots. Each trial has a 180-second
limit, and each low-level call in SCTD and MMD samples 64 candidates.

Both XECBS systems retain inter-agent constraint guidance and post-sampling smoothing.
SCTD disables static-obstacle and smoothness cost guidance, while MMD retains
these terms.

Table~\ref{tab:multiagent-success} reports team success: the percentage
of tasks solved within 180 seconds with feasible trajectories for all
robots and no detected inter-agent conflicts.
At every reported team size, SCTD has higher team success than the
MMD baseline. At the shared sizes of 3, 6, and 9 robots, it also
exceeds SMD.


Figure~\ref{fig:multiagent-win-by-n} compares selected multi-robot
plans for teams of 6, 9, and 15 robots. In these examples, MMD
produces trajectories with obstacle collisions, whereas SCTD
routes the highlighted robots around the obstacles. The
nine-robot example shows a particularly clear routing difference:
SCTD takes the highlighted robot around the opposite side of the
central obstacle. These cases illustrate obstacle-avoidance
differences within the shared XECBS coordination framework,
complementing the success results in
Table~\ref{tab:multiagent-success}.

\FloatBarrier
\section{Conclusion}
We presented SCTD, a spatially map-conditioned trajectory generator.
Multiple position-aware occupancy tokens allow trajectory features to access different workspace regions through cross-attention throughout denoising. On held-out maps, SCTD achieves higher single-agent success than the MPD baseline without sampling time obstacle cost guidance, with the largest gains on hard-detour tasks. The conditioning-design ablation further shows higher success with spatial tokens than with a global map embedding. Integrating SCTD into XECBS supports multi-robot motion planning.
This combination of spatial map conditioning and high-level conflict-based coordination enables one model trained on single-robot demonstrations to serve different robots, workspace layouts, and team sizes, while inter-agent guidance addresses coordination constraints during sampling and replanning.
This integration of SCTD into XECBS achieves higher team success than MMD at all evaluated team sizes and than SMD at the shared sizes of 3, 6, and 9 robots. 

\FloatBarrier

\bibliographystyle{plainnat}
\bibliography{references}

\clearpage
\appendix

\section{Implementation Details}
\label{sec:implementation-details}

\paragraph{Hardware and software.}
All experiments ran on NVIDIA RTX 6000 Ada Generation GPUs (48 GB) with Intel Xeon Gold 6438Y+
CPUs, using Python 3.8.20, PyTorch 2.0.0, and CUDA 11.8. Our implementation builds on the
official code of MMD~\cite{shaoul2025mmd}, which in turn builds on MPD~\cite{carvalho2025mpd}.

\paragraph{Trajectory representation.}
A trajectory has 64 waypoints with state $(x,y,\dot x,\dot y)$. Each channel is scaled to $[-1,1]$
with limits fitted on the training split: positions use their minima and maxima, and both velocity
channels share the 99.9th percentile of absolute velocity, above which values saturate. Start and
goal states have zero velocity. Generated trajectories are converted back to world coordinates
once, before evaluation.

\paragraph{Network architecture.}
Table~\ref{tab:architecture} lists the layer dimensions. Each convolution in the occupancy encoder
is followed by group normalization and a SiLU activation. Cross-attention uses two heads of
dimension 32. The SDF pretraining head receives the bilinearly interpolated 64-dimensional stem
feature concatenated with the query coordinates and applies three hidden layers of width 64 with
Mish activations and a scalar output. It is discarded after pretraining.

\begin{table}[htb]
\centering
\small
\caption{Layer dimensions of SCTD. Convolutional tuples give
(output channels, kernel size, stride, padding);
$B$ denotes batch size.}
\label{tab:architecture}
\resizebox{\linewidth}{!}{%
\begin{tabular}{lll}
\toprule
Module & Layer & Dimensions \\
\midrule
\multicolumn{3}{l}{\textbf{Occupancy encoder}} \\
& Input & $[B,1,64,64]$ \\
& Conv2d & $(32,3,2,1)\rightarrow[B,32,32,32]$ \\
& Conv2d & $(64,3,2,1)\rightarrow[B,64,16,16]$ \\
\midrule
\multicolumn{3}{l}{\textbf{Spatial map tokens}} \\
& Flatten spatial grid & $[B,256,64]$ \\
& Learned positional embedding & $[1,256,64]$ \\
& Context output & $[B,256,64]$ \\
\midrule
\multicolumn{3}{l}{\textbf{Temporal U-Net}} \\
& Trajectory input & $[B,64,4]$ \\
& Down stages & $(4\rightarrow32,64),\ (32\rightarrow64,32),\ (64\rightarrow128,16)$ \\
& Middle stage & $[B,128,16]$ \\
& Up stages & $128\rightarrow64\rightarrow32$ channels; horizons $16\rightarrow32\rightarrow64$ \\
& Output projection & $[B,64,4]$ predicted noise \\
\midrule
\multicolumn{3}{l}{\textbf{Conditioning}} \\
& Cross-attention context & 256 keys/values of dimension 64 \\
& Attention locations & every down, middle, and up block \\
& Diffusion-time conditioning & additive channel-wise bias in every residual block \\
& Hard endpoint conditions & trajectory rows $\{0,63\}$ \\
\bottomrule
\end{tabular}}
\end{table}
\clearpage
\paragraph{Training and sampling.}
Table~\ref{tab:config} lists the hyperparameters of both training stages and of sampling.

\begin{table}[t]
\centering
\small
\caption{Training and sampling hyperparameters of SCTD.}
\label{tab:config}
\resizebox{\linewidth}{!}{%
\begin{tabular}{ll}
\toprule
Setting & Value \\
\midrule
\multicolumn{2}{l}{\textbf{Stage 1: SDF pretraining}} \\
Maps & $45{,}450$ \\
Query points per map & $1{,}024$, uniform over the workspace \\
Target & signed distance clipped to $[-0.2,0.2]$ \\
Optimizer & Adam, learning rate $10^{-3}$ \\
Batch size / epochs & $32$ / $4$ \\
\midrule
\multicolumn{2}{l}{\textbf{Stage 2: diffusion training}} \\
Demonstrations & $44{,}984$ on $4{,}500$ maps \\
Optimizer & Adam, constant learning rate $10^{-4}$ \\
Batch size & $32$ \\
Training steps & $1{,}000{,}000$ \\
Gradient clipping & global norm $1.0$ \\
Reported weights & EMA, decay $0.995$, updated every 10 steps from step $1{,}000$ \\
\midrule
\multicolumn{2}{l}{\textbf{Sampling}} \\
Sampler & DDPM, $25$ steps, last step noiseless \\
Variance schedule & exponential, $\beta_{\mathrm{start}}=10^{-4}$, $\beta_{\mathrm{end}}=1.0$ \\
Extra reconstruction & one noiseless pass at diffusion step zero \\
Reverse-noise multiplier & $0.5$ times the posterior standard deviation \\
Clean-trajectory estimate & clamped to $[-1,1]$ at each step \\
Candidates per task & $64$ \\
\bottomrule
\end{tabular}}
\end{table}

\FloatBarrier
\section{Dataset Generation}
\label{sec:dataset-generation}

\paragraph{Maps.}
Each map is a $2\times2$ workspace containing four to six obstacles: disks with radius in
$[0.20,0.32]$ and axis-aligned rectangles with half-extents in $[0.16,0.30]$. A fraction of
obstacles is placed adjacent to a neighbour to form compound shapes that force detours; the
others keep a gap of at least $0.11$, wider than the robot diameter. Maps are generated
deterministically from their index and rasterized to a $64\times64$ occupancy grid.

\paragraph{Tasks.}
Start and goal positions are sampled with obstacle clearance at least $0.16$, at least $0.15$ apart,
and in the same connected free region of a $128\times128$ grid.

\paragraph{Expert trajectories.}
For each of 30 candidate tasks per training map, a single RRTConnect--GPMP2 run produces a
64-waypoint trajectory (RRTConnect initialization followed by 500 GPMP2 optimization iterations).
We discard trajectories that collide, and for straight-clear tasks those whose arc length
exceeds $1.10$ times the start--goal distance. From the remaining trajectories, each map keeps at
most ten, aiming for one straight-clear and nine obstacle-dependent examples while preserving the
available ratio of mild to hard-detour tasks. The resulting diffusion-training set contains
44,984 demonstrations on 4,500 maps.

\paragraph{Encoder-pretraining maps.}
SDF pretraining uses 45,450 maps from the same generator: 450 of the diffusion-training maps and
45,000 additional maps. Pretraining needs no demonstrations.

\paragraph{Evaluation maps.}
Training, held-out evaluation, and test maps are disjoint by index. The single-agent benchmark
consists of 500 test maps with 30 tasks each, taken in sampling order among candidates that pass
a $512\times512$ four-connected reachability check for the robot. The map-dependence control in
Appendix~\ref{sec:map-dependence-control} uses a separate set of 500 held-out evaluation maps.
\section{Multi-Robot Planning Details}
\label{sec:multi-robot-procedure}

Algorithm~\ref{alg:sctd-xecbs} summarizes how SCTD is used within XECBS~\cite{shaoul2025mmd}.
Each node stores conflict constraints, a batch of candidate trajectories for each robot, and one
selected trajectory per robot. $\operatorname{Plan}(i,n,\mathcal B_{\mathrm{prev}})$ samples a new
batch for robot $i$ from SCTD, conditioned on the occupancy map and the robot's endpoints and guided
by its conflict constraints and by soft constraints derived from the other robots' selected
trajectories in $n$. With an empty previous batch, sampling starts from Gaussian noise; otherwise,
the previous batch is re-noised for 3 steps and denoised for 3 steps. A call fails if no candidate
is free of obstacle collisions. At the root, each robot selects its lowest-cost collision-free
candidate; after replanning, the collision-free candidate with the fewest inter-robot conflicts.

\paragraph{Parameters.}
Each call samples 64 candidates; initial planning uses 25 denoising steps. As in MMD, guidance is applied during the
final half of the reverse process with 20 gradient steps per denoising step. Conflict constraints
are spheres of radius $0.12$ (weight $2\times10^{-1}$), and soft constraints use weight
$2\times10^{-2}$. SCTD sets the static-obstacle and smoothness guidance weights to zero; the MMD
baseline uses its defaults of $2\times10^{-2}$ and $8\times10^{-2}$. Returned trajectories are
smoothed with a Savitzky--Golay filter (window 10, order 2).

\begin{algorithm}[htb]
\small
\caption{SCTD within XECBS, adapted from
Shaoul et al.~\cite{shaoul2025mmd}}
\label{alg:sctd-xecbs}
\begin{algorithmic}[1]
\Statex \textbf{Input:} occupancy map $M$,
endpoint states $\{(s_i^{\mathrm{s}},s_i^{\mathrm{g}})\}_{i=1}^{N}$,
shared SCTD model, runtime limit
\Statex \textbf{Output:} selected trajectories with no detected
inter-agent conflicts, or failure

\State $n_0 \gets$ new node with empty constraints,
batches, and selected trajectories
\For{$i=1,\ldots,N$}
  \If{$\operatorname{Plan}(i,n_0,\varnothing)$ fails}
    \State \Return failure
  \EndIf
\EndFor
\State $\mathcal Q \gets \{n_0\}$
\While{$\mathcal Q\neq\varnothing$ and time remains}
  \State Remove from $\mathcal Q$ a node $n$ with the
  fewest inter-agent conflicts
  \If{the selected trajectories in $n$ have no detected conflict}
    \State \Return $n.\boldsymbol{\tau}$
  \EndIf
  \State $c \gets$ a detected conflict between robots $i$ and $j$
  \For{$k\in\{i,j\}$}
    \State $n' \gets$ copy of $n$
    \State Add the conflict constraint for robot $k$
    derived from $c$ to $n'.C_k$
    \If{$\operatorname{Plan}(k,n',n.\mathcal B_k)$ succeeds}
      \State Insert $n'$ into $\mathcal Q$
    \EndIf
  \EndFor
\EndWhile
\State \Return failure
\end{algorithmic}
\end{algorithm}
\section{Evaluation Details}
\label{sec:evaluation-protocol}

\paragraph{Single-agent feasibility.}
A trajectory is feasible if, at every waypoint and at three interpolated points per segment, the
robot clears every obstacle by its radius and its centre stays inside the workspace. For
candidate budget $J$, let $F_{mtj}$ indicate whether draw $j$ for task $t$ on map $m$ is feasible,
and let $T_m$ be the map's tasks in the reported group. Success@$J$ is
\begin{equation*}
  \widehat S_J=\frac{1}{|\mathcal M_*|}\sum_{m\in\mathcal M_*}
  \frac{1}{|T_m|}\sum_{t\in T_m}
  \mathbf 1\!\left\{\max_{1\leq j\leq J}F_{mtj}=1\right\},
\end{equation*}
\begin{equation*}
  \qquad \mathcal M_*=\{m:|T_m|>0\}.
\end{equation*}

\paragraph{Task groups.}
A task is straight-clear if the straight start--goal segment clears all obstacles by the robot
radius, and obstacle-dependent otherwise. An obstacle-dependent task is hard-detour if its shortest
8-connected path on the $64\times64$ grid is at least $1.2$ times the start--goal distance, and mild
otherwise. Tasks that are reachable at $512\times512$ but disconnected at $64\times64$ are kept and
counted as mild.

\paragraph{Trajectory quality.}
Path length is the sum of distances between consecutive waypoints. Smoothness is the maximum local
turn angle: the largest angle between consecutive chords spanning a path distance of one robot
diameter ($w=0.1$). Quality comparisons use tasks for which every compared method produces at least
one feasible trajectory; metrics are averaged over feasible draws within a task, then over tasks
within a map, and then equally over maps.

\paragraph{Multi-robot success.}
A trial succeeds if XECBS returns a solution within 180 seconds, no two robots come closer than
$0.105$ at any synchronized waypoint, no robot's smoothed trajectory collides with an obstacle
under MMD's grid-based collision check, and every robot's smoothed trajectory starts and ends
within $0.05$ of its endpoints. SMD trajectories are scored with the same checks.

\paragraph{Single-agent baselines.}
MPD uses MMD's released Empty prior (\texttt{EnvEmpty2D-RobotPlanarDisk}) with its default
SDF-based obstacle and smoothness guidance computed on the evaluated map, followed by MMD's
Savitzky--Golay smoothing; it is scored after smoothing, and SCTD on its raw output. We use the
Empty prior because MMD's other priors encode application-specific motion patterns that are absent
from our maps. A* plans on a 4-connected grid with spacing $0.1$, using exact obstacle geometry;
each endpoint is connected by straight segments to the free corners of its grid cell. A* returns
one unsmoothed route per task.

\paragraph{Multi-robot baselines.}
MMD runs XECBS with the Empty prior and its default guidance weights. SMD~\cite{liang2025smd} uses its
released composite models, which exist for teams of 3, 6, and 9 robots. Its projection step handles
only circular obstacles, so each rectangle is replaced by a disk cover with maximum boundary error
$0.01$, shared by guidance and projection. The projection uses robot radius $0.055$, so its accepted
solutions satisfy our separation and clearance thresholds, and is initialized with the model's own
sample. SMD uses 64 samples and 25 denoising steps, as MMD and SCTD do. The 180-second limit is
enforced externally, and BLAS is limited to one thread per trial. We did not run the
disk-cover-resolution and A*-initialization sensitivity checks that we had planned.

\section{Additional Results}
\label{app:supp-results}

\subsection{Map-dependence control}
\label{sec:map-dependence-control}

To test whether SCTD uses its map input, we replace the supplied occupancy map with a different map
while keeping the start and goal fixed, and score the result against the original obstacles. We use
obstacle-dependent tasks solved by the expert planner on the 500 evaluation maps held out from
training. For each task, we select up to 15 replacement maps with clear, connected endpoints using a
fixed seed. We then keep only those that block the straight segment between the start and goal. We
exclude 39 tasks with no remaining replacement. For each task, we generate 64 samples per replacement
map, sharing sampling noise across its replacements.

\begin{figure}[p]
  \centering
  \includegraphics[
    width=\columnwidth,
    height=0.82\textheight,
    keepaspectratio
  ]{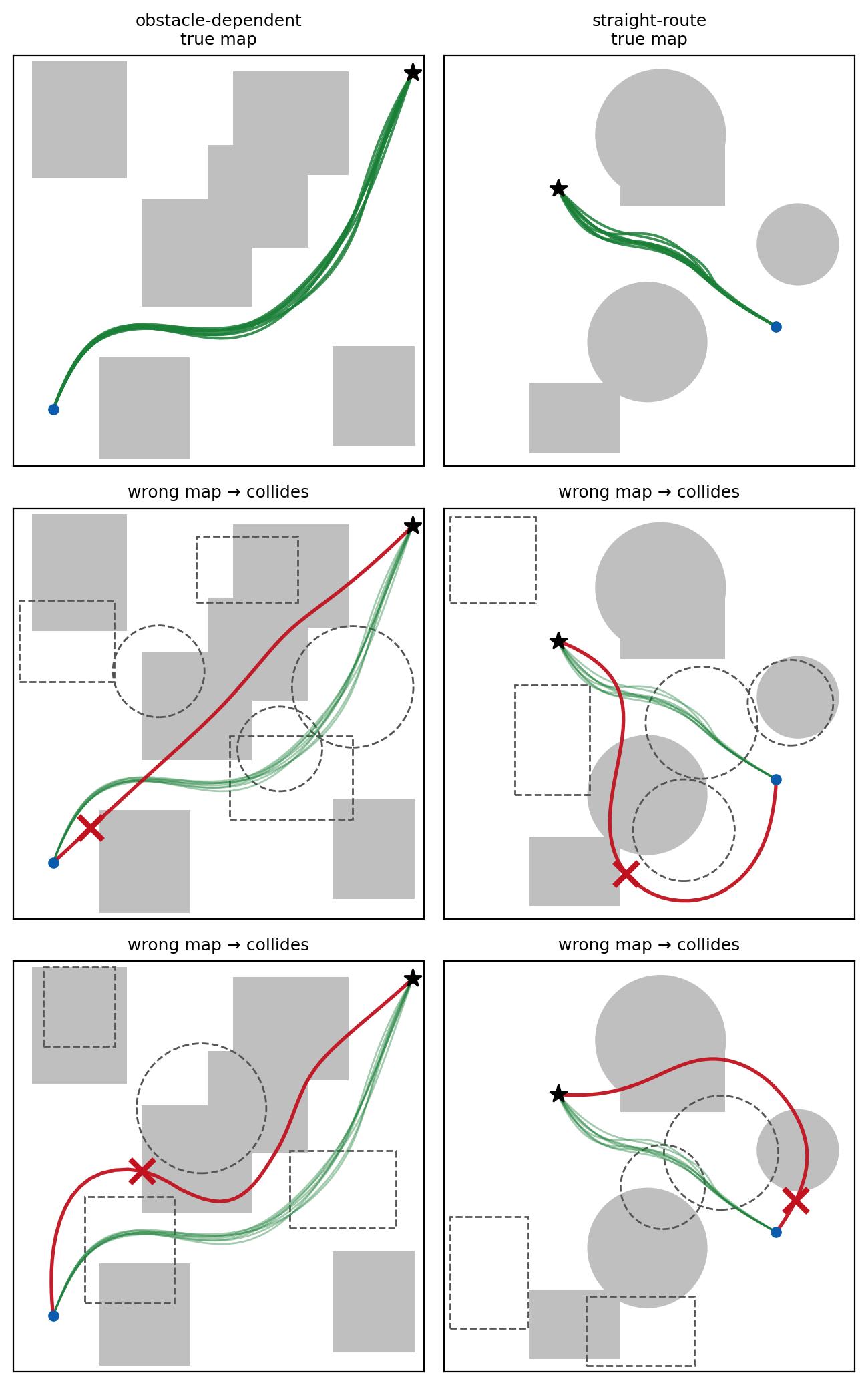}
  \caption{Same-noise map replacement for an obstacle-dependent task (left) and a
  straight-route task (right). The top row shows collision-free draws on the true map. In the
  lower rows, solid grey obstacles are the true walls used for scoring, dashed outlines show the
  supplied replacement map, faint green curves repeat the true-map paths, and the bold red curve
  is the draw produced from identical noise under the replacement; each collides.}
  \label{fig:wrong-map}
\end{figure}

\begin{figure*}[p]
  \centering
  \includegraphics[width=\linewidth]{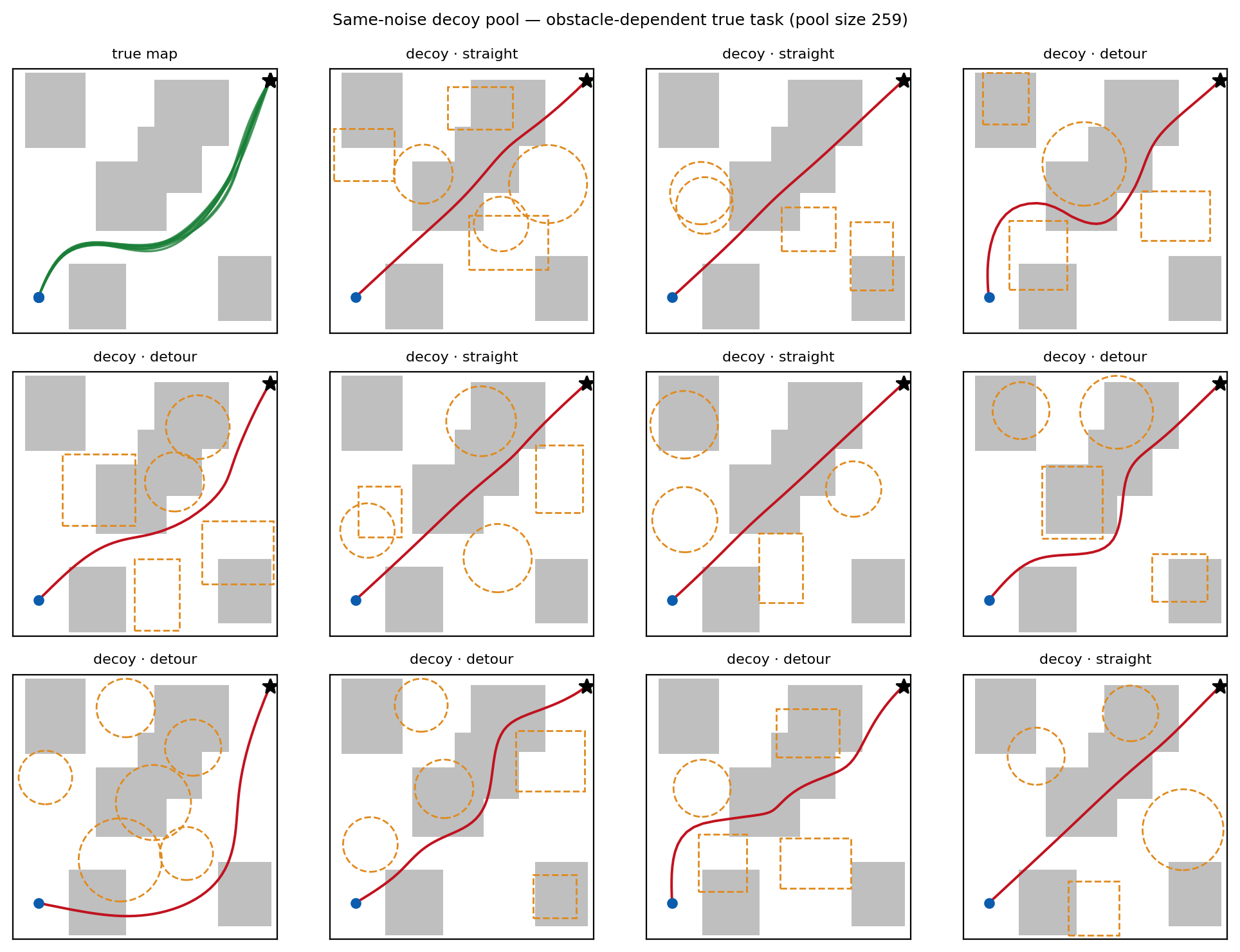}
  \caption{All replacement maps for one obstacle-dependent task. The top-left panel shows the true
  map and true-map draws. Other panels show the true walls (solid grey), the replacement map
  (dashed orange), and the same-noise draw, green if collision-free and red if it collides.}
  \label{fig:pool-950}
\end{figure*}

Over 7,336 tasks and 84,373 replacement comparisons, the collision-free rate is 87.3\% with the
correct map and 7.6\% with a replacement map. Within each original map, we average success with the
correct map over tasks and success with replacement maps over replacement comparisons. We then
average both rates equally across original maps. The difference is 79.7 percentage points, with a 95\%
bootstrap interval of $[78.4, 80.9]$, computed from 10,000 resamples over both original and replacement
maps. Figures~\ref{fig:wrong-map} and~\ref{fig:pool-950} show selected qualitative examples from a
separate pool of replacement maps.

\subsection{Location of large turns}
\label{sec:large-turn-location}
Among feasible MPD paths whose maximum local turn angle is at least $90^\circ$, the maximum lies in
the final 10\% of the path in 62.6\% of cases overall, 96.4\% on mild obstacle-dependent tasks,
and 97.3\% on hard-detour tasks. For SCTD the overall share is 4.0\%, and only 1.5\% of its
feasible paths contain a turn of at least $90^\circ$.

\subsection{Qualitative results}

\begin{figure*}[htb]
  \centering
  \includegraphics[width=\linewidth]{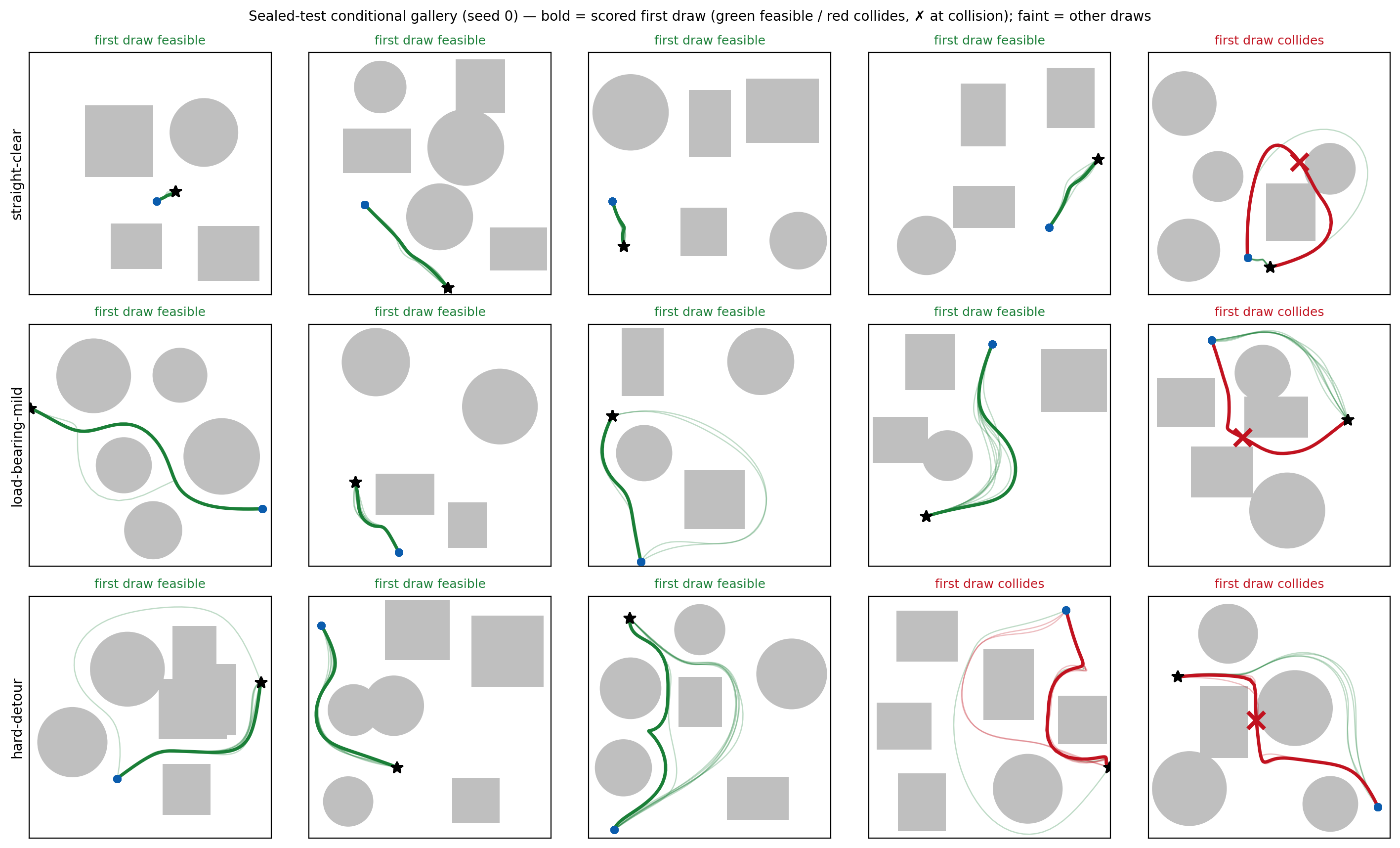}
  \caption{SCTD on test tasks. Rows are straight-clear, mild obstacle-dependent, and hard-detour
  tasks. Each panel overlays several of the 64 raw draws; the bold curve is the first draw, and
  green and red mark collision-free and colliding draws. Each row has a fixed number of
  first-draw failures (one, one, and two), filled at random with seed 0, so the rows do not
  reflect success rates.}
  \label{fig:results-gallery}
\end{figure*}

\paragraph{Figure selection.}
For the main-text trajectory figure, each task is drawn at random (seed 1) from test tasks in its
group whose first SCTD draw is feasible. Figure~\ref{fig:results-gallery} shows additional test
tasks.

\FloatBarrier
\subsection{Unseen obstacles in MMD environments}
\label{app:ood-mmd-envs}

We add obstacles to MMD's Conveyor and Highways environments on the routes followed by MPD's
training demonstrations. All added obstacles are axis-aligned rounded boxes, every scene remains
connected, the workspace is $[-1,1]^2$, and the robot radius is $0.05$.

\paragraph{Scenes.}
In Conveyor, OOD-1 adds a 0.2-unit bump to the middle bar of the top or bottom corridor, leaving a
0.15-unit gap (two maps). OOD-2 adds a chicane of 0.1-unit bumps in either corridor, with
0.15-unit gaps (two maps). OOD-3 and OOD-4 apply the OOD-1 and OOD-2 changes to both corridors (one
map each). OOD-5 adds a 0.2-unit plug closing either corridor (two maps). In Highways, OOD-1 adds a
$0.1\times0.2$ block in each of the four lanes in turn, leaving gaps of 0.25 and 0.15; OOD-2 centres
the same block, leaving two 0.20 gaps; OOD-3 uses a $0.2\times0.2$ block, leaving two 0.15 gaps;
OOD-4 adds a $0.6\times0.2$ bar to OOD-3, closing the counter-clockwise route.

\paragraph{Protocol.}
Each map has 200 tasks from MMD's random endpoint sampler (obstacle clearance 0.16), kept if
reachable. Endpoints are sampled independently on each map, while the sampling-noise seed is fixed
across scenes. Rates are averaged equally across maps, using the feasibility check and 64-sample
protocol of Appendix~\ref{sec:evaluation-protocol}. MPD uses MMD's released checkpoint for each
environment with SDF-based guidance and is scored after smoothing; SCTD, which was not trained on
either environment, receives the scene's $64\times64$ occupancy map and is scored on its raw
output. The detour subset contains tasks whose shortest path is longer than in the Original
scene. Main-text scenes were chosen after the results were known: per environment, among scenes
where SCTD's Success@1 exceeds MPD's by at least 10 points, up to three with the largest
Success@64 gap. Table~\ref{tab:ood-mmd-envs-full} reports all scenes.

\paragraph{Illustrations.}
The main-text illustrations show the first draw for one task per scene whose straight start--goal
segment is blocked; a red cross marks the first collision. For scenes with added obstacles, the task
is the one closest to the scene's mean feasible-draw counts among tasks where SCTD has more
feasible draws than MPD.

\begin{table*}[t]
  \caption{Single-agent success (\%) in MMD's Conveyor and Highways environments. ``--'' marks an empty detour subset.}
  \label{tab:ood-mmd-envs-full}
  \centering
  \small
  \setlength{\tabcolsep}{3pt}
  \begin{tabular}{llcrrrrrrrrr}
    \toprule
    Environment & Scene & Maps & \multicolumn{4}{c}{All tasks} & \multicolumn{5}{c}{Detour subset} \\
    \cmidrule(lr){4-7}\cmidrule(lr){8-12}
    & & & \multicolumn{2}{c}{MPD} & \multicolumn{2}{c}{SCTD} & $n$ & \multicolumn{2}{c}{MPD} & \multicolumn{2}{c}{SCTD} \\
    \cmidrule(lr){4-5}\cmidrule(lr){6-7}\cmidrule(lr){9-10}\cmidrule(lr){11-12}
    & & & @1 & @64 & @1 & @64 & & @1 & @64 & @1 & @64 \\
    \midrule
    Conveyor & Original & 1 & \textbf{98.5} & \textbf{100.0} & 65.5 & \textbf{100.0} & -- & -- & -- & -- & -- \\
             & OOD-1    & 2 & 51.2 & \textbf{100.0} & \textbf{60.8} & \textbf{100.0} & -- & -- & -- & -- & -- \\
             & OOD-2    & 2 & 48.0 & 98.8  & \textbf{63.7} & \textbf{100.0} & 8 & \textbf{62.5} & \textbf{100.0} & 50.0 & \textbf{100.0} \\
             & OOD-3    & 1 & 4.5  & 71.0  & \textbf{62.0} & \textbf{99.5}  & -- & -- & -- & -- & -- \\
             & OOD-4    & 1 & 0.0  & 7.0   & \textbf{73.0} & \textbf{100.0} & 8 & 0.0  & 0.0   & \textbf{50.0} & \textbf{100.0} \\
             & OOD-5    & 2 & 47.5 & 98.5  & \textbf{64.0} & \textbf{99.8}  & 18 & 27.8 & \textbf{100.0} & \textbf{33.3} & \textbf{100.0} \\
    \midrule
    Highways & Original & 1 & \textbf{100.0} & \textbf{100.0} & \textbf{100.0} & \textbf{100.0} & -- & -- & -- & -- & -- \\
             & OOD-1    & 4 & 88.5  & \textbf{100.0} & \textbf{93.6}  & \textbf{100.0} & 37 & \textbf{86.5} & \textbf{100.0} & 67.6 & \textbf{100.0} \\
             & OOD-2    & 4 & \textbf{99.8}  & \textbf{100.0} & 90.5  & \textbf{100.0} & 46 & \textbf{97.8} & \textbf{100.0} & 58.7 & \textbf{100.0} \\
             & OOD-3    & 4 & 51.6  & 89.4  & \textbf{87.1}  & \textbf{100.0} & 50 & 56.0 & \textbf{100.0} & \textbf{64.0} & \textbf{100.0} \\
             & OOD-4    & 4 & 44.6  & 49.5  & \textbf{82.4}  & \textbf{100.0} & 175 & \textbf{46.9} & 46.9  & 39.4 & \textbf{100.0} \\
    \bottomrule
  \end{tabular}
\end{table*}

\end{document}